\documentclass[a4paper,fleqn]{cas-dc}

\usepackage[authoryear,longnamesfirst]{natbib}

\usepackage{dirtytalk} 
\usepackage{gensymb} 
\usepackage{pdflscape}

\def\tsc#1{\csdef{#1}{\textsc{\lowercase{#1}}\xspace}}
\tsc{WGM}
\tsc{QE}
\tsc{EP}
\tsc{PMS}
\tsc{BEC}
\tsc{DE}

\begin{document}
\let\WriteBookmarks\relax
\def\floatpagepagefraction{1}
\def\textpagefraction{.001}

\shorttitle{Object-centric visual reasoning}

\shortauthors{Puebla \& Bowers}

\title [mode = title]{Visual Reasoning in Object-Centric Deep Neural Networks: A Comparative Cognition Approach}                      



%
\author[1]{Guillermo Puebla}[
  orcid=0000-0001-7002-7776]

\cormark[1]


\ead{pueblaramirezg@gmail.com}


\credit{Conceptualization, Formal analysis, Software, Writing - Original draft preparation}

\affiliation[1]{organization={Instituto de Alta Investigación, Universidad de Tarapacá},
    addressline={Casilla 7D}, 
    city={Arica},
    citysep={}, 
    postcode={1000000}, 
    country={Chile}}

\author[2]{Jeffrey S. Bowers}[%
   orcid=0000-0001-9558-5010
   ]

\credit{Conceptualization, Funding acquisition, Writing - Review \& Editing}

\affiliation[2]{organization={School of Psychological Science, University of Bristol},
    addressline={12a Priory Road}, 
    city={Bristol},
    citysep={}, 
    postcode={BS8 1TU}, 
    country={UK}}

\cortext[cor1]{Corresponding author}



\begin{abstract}
Achieving visual reasoning is a long-term goal of artificial intelligence. In the last decade, several studies have applied deep neural networks (DNNs) to the task of learning visual relations from images, with modest results in terms of generalization of the relations learned. However, in recent years, object-centric representation learning has been put forward as a way to achieve visual reasoning within the deep learning framework. Object-centric models attempt to model input scenes as compositions of objects and relations between them. To this end, these models use several kinds of attention mechanisms to segregate the individual objects in a scene from the background and from other objects. In this work we tested relation learning and generalization in several object-centric models, as well as a ResNet-50 baseline. In contrast to previous research, which has focused heavily in the same-different task in order to asses relational reasoning in DNNs, we use a set of tasks ---with varying degrees of difficulty--- derived from the comparative cognition literature. Our results show that object-centric models are able to segregate the different objects in a scene, even in many out-of-distribution cases. In our simpler tasks, this improves their capacity to learn and generalize visual relations in comparison to the ResNet-50 baseline. However, object-centric models still struggle in our more difficult tasks and conditions. We conclude that abstract visual reasoning remains an open challenge for DNNs, including object-centric models.

\end{abstract}




\maketitle

\section{Introduction}

The human visual system produces a rich and dynamic parsing of the visual world into discrete objects and relations among them \citep{biederman1981semantics}. What underlies this fact is an ability to represent relations between entities in way that can be generalized across specific contexts and objects \citep[for reviews, see][]{green2004relational,hafri2021perception}. Relational reasoning has been identified as a mayor contributor to humans' unique cognitive abilities \citep{gentner2010bootstrapping,penn2008darwin}. In fact, relational processing pervades human cognition, from object recognition \citep{biederman1987recognition} to metaphor comprehension \citep{holyoak2018metaphor}. Consequently, in recent years researchers have developed several architectural innovations in order to achieve relational reasoning in visual processing deep neural networks \citep[DNNs; for reviews, see][]{greff2020binding,ricci2021same,malkinski2023review}.

In this work we will focus on models that fall under the umbrella of \emph{object-centric} representation learning methods \citep{dittadi2022generalization}. These DNNs have specific inductive biases targeted towards segregating the individual objects in a scene from the background and from other objects. Object-centric representations are hypothesized to be more robust than standard more holistic representations to low-level perturbations \citep[e.g., changes in background;][]{greff2020binding}, to enable systematic generalization \citep{locatello2020object} and to be the basis for the construction of world models from sensory data \citep{kipf2019contrastive}.

Although a wide range of DNN architectures can be classified as object-centric, a common attribute across these models is the use of attention mechanisms that isolate individual objects. For most models, this is done through what has been termed affinity-based attention \citep{adeli2023affinity} or perceptual grouping \citep{mehrani5self}. This process can be described as a bottom-up clustering of the visual input based on perceptual features such as color, texture and position \citep{mehrani5self,locatello2020object}.

\begin{figure}[t]
    \centering
    \includegraphics[width=1.0\linewidth]{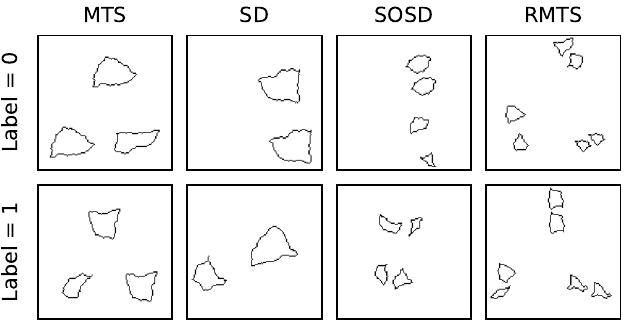}
    \caption{Visual reasoning tasks. MTS: match-to-sample; SD: same-different; SOSD: second-order same-different; RMTS: relational match-to-sample. See text for details.}
    \label{fig:tasks}
\end{figure}

Importantly, previous research has shown that object segregation improves out-of-distribution generalization in some visual tasks. For example, \citet{dittadi2022generalization} found that the segmentation performance of slot-based DNNs (a type of object-centric models, see section \ref{section:models}) is generally robust to local distribution shifts such as presenting a scene with a single object with an unseen color or texture.  (However, the usefulness of object-centric representations for downstream tasks was severely hindered by more global shifts like cropping.) Closely related to the current work, \citet{puebla2022can} assessed the impact of object segregation on the same-different task. This task consist of classifying pairs of objects as examples of the categories \say{same} or \say{different} (see Fig. \ref{fig:tasks} second column). \citet{puebla2022can} simulated object segregation by feeding individual objects to different channels of a Siamese Network \citep{bromley1993signature}. This partially improved same-different discrimination in out-of-distribution samples (i.e., images perceptually dissimilar to the training data but that followed the same classification rule) in comparison to standard ResNet \citep{he2016deep} models. Note, however, that overall generalization performance was far from optimal.

Previous research on the visual reasoning capabilities of DNNs has focused heavily on the same-different task \citep{adeli2023brain,baker2023configural,funke2021five,kim2018not,messina2021solving,messina2022recurrent,puebla2022can,ricci2021same,stabinger2021arguments,tartaglini2023deep,vaishnav2022understanding,webb2023systematic,webb2021emergent}. This is due to the fact that the concept of sameness is considered to be fundamental to human thought \citep{hochmann2021editorial}, develops early in human infants \citep{hespos2021origins}, and it is more sophisticated in humans in comparison to other species \citep{gentner2021learning}. Furthermore, as putative models of the human visual system, DNNs should be able to support not only object recognition, but also visual reasoning \citep{bowers2023deep}. In keeping with this line of research, in this work we studied learning and generalization of the sameness relation in DNNs. However, in contrast with previous work, we used a family of tasks designed to assess sameness understanding (including the standard same-different task) derived from the comparative cognition literature. As stated previously, our focus will be on object-centric models that have been claimed to perform visual reasoning.

\section{Tasks}

Our tasks are based on David Premack's \citeyearpar{premack1983codes} influential work on primate reasoning, where he classifies same-different perception tasks into three classes. In the \emph{match-to-same} task (MTS; Fig. \ref{fig:tasks} first column) there is a base shape at the top-center of the image and two candidate shapes at the bottom-left and bottom-right. One of these candidates is the same as the base (up to translation) and the goal is to identify which candidate matches the base. A given image is labeled as 0 if the match is to the bottom-left and as 1 if the the match is to the bottom-right. As described earlier, the \emph{same-different} task (SD; Fig. \ref{fig:tasks} second column) involves classifying images of pairs of objects as \say{same} (label 0) or \say{different} (label 1). In the last task defined by Premack, the \emph{relational match-to-same} task (RMTS; Fig. \ref{fig:tasks} fourth column), there is a base pair of shapes at the top-center of the image exemplifying the relation \say{same} or \say{different} and two candidate pairs of shapes at the bottom-left and bottom-right. One of these candidate pairs exemplifies the same relation as the base pair and the goal is to identify which candidate pair matches the base's relation. As in the MTS task, a given image is labeled as 0 if the match is to the bottom-left and as 1 if the the match is to the bottom-right. Finally, because of the higher number of objects of the RMTS task in comparison to SD and MTS, we created a new task designed to impose similar representational demands to RMTS but with fewer objects. The \emph{second-order same-different} task (SOSD, Fig. \ref{fig:tasks} third column) does this by asking models to classify whether two pairs of objects, one at the center top and the other at the center bottom, exemplify a single relation (i.e., \say{same-same} or \say{different-different}; labeled as 1) or not (i.e., \say{same-different} or \say{different-same}; labeled as 0). 

Comparative research has shown that the tasks identified by Premack follow a hierarchy of difficulty, with fewer and fewer species being able to pass the MTS, SD and RMTS tasks, respectively \citep[for a review, see][]{gentner2021learning}. This general trend also applies to human infants, with each task being able to be passed by progressively older children \citep[for reviews, see][]{gentner2021learning,hespos2021origins,hochmann2021asymmetry,shivaram2023children}. Overall, these tasks seem to have different computational demands for the subject. While the MTS task can be passed through the use of an general sense of similarity, the SD task requires the decision maker to classify the relation between the objects in the image. In contrast, the RMTS task requires to compare the relations between the different pairs of objects (which at least in principle, requires registers that can temporarily store the relations at play).

In order to assess whether the models were able to generalize their learning on these tasks, we followed the out-of-distribution testing protocol of \citet{puebla2022can}. For each task we trained the models in one \emph{Original} dataset composed of shapes taken from the synthetic visual reasoning test \citep[SVRT;][]{fleuret2011comparing} and tested them on the test split of this dataset plus 13 out-of-distribution datasets that followed the same classification rule but whose objects had different perceptual features (see section \ref{section:simulations} for details). Furthermore, to evaluate whether these models indeed learned object-centric representations, we generated model attribution visualizations and examined the degree to which the models focused on individual objects instead of the whole input images.

\begin{figure*}[!ht]
    \centering
    \includegraphics[width=0.95\linewidth]{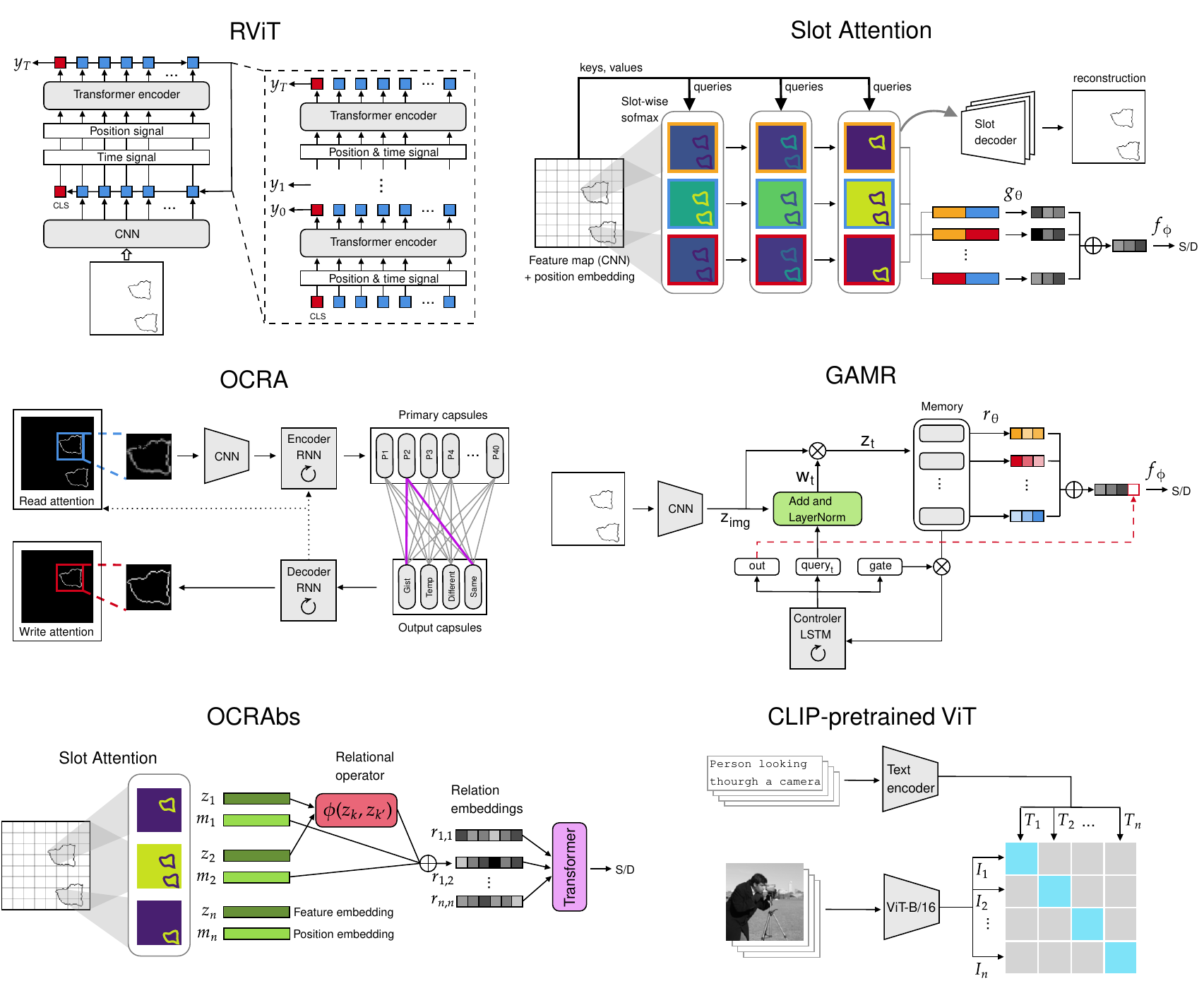}
    \caption{Models tested.}
    \label{fig:models}
\end{figure*}

\section{Models}\label{section:models}

We used ResNet50 \citep{he2016deep} as our deep CNN baseline. We compared the generalization performance of this baseline with six object-centric models, which we briefly describe next. For all these models we used the hyperparameter settings available in their respective publications, adapting them when necessary to improve validation performance. All the scripts necessary to train and test our models are available at the the article's repository: \url{https://github.com/GuillermoPuebla/object-centric-reasoning}.

The first two models are vision transformers \citep[ViT,][]{dosovitskiy2020image}. Although not specifically designed for object segmentation, recent research has shown that, when trained in a self-supervised manner, standard ViTs tend to group object features together \citep{adeli2023affinity,caron2021emerging,oquab2023dinov2}, which lead us to include ViTs in our comparisons. The first model was the Recurrent Vision Transformer \citep[RViT,][see Fig. \ref{fig:models} top left]{messina2022recurrent}. This architecture applies a ViT encoder recurrently, that is, the model takes as input its own output for several processing steps (see Fig. \ref{fig:models} top left). \citet{messina2022recurrent} showed that RViT was more efficient at learning the SD task than several CNNs (as well as a standard ViT trained from scrath that was not able to learn this task). They also showed that the RViT attention maps tended to focus of the relevant objects across time steps. From these results, \citet{messina2022recurrent} conclude that recurrent connections are a key component to solve the SD task. Note, however, that \citet{messina2022recurrent} did not test the generalization performance of the RViT in out-of-distribution samples of the SD task. 

The second model was a CLIP-pretrained ViT-B/16 \citep[][heceforth, CLIP-ViT]{radford2021learning}. CLIP models are pretrained with a text-image contrastive loss that maximizes the the cosine similarity between an image embedding and its matching natural language description (see Fig. \ref{fig:models} bottom-right). Recently, \citet{tartaglini2023deep} showed that, when trained in a custom SD dataset very similar to the one used in the current work, this model was able to generalize same-different classification to seven out-of-distribution datasets taken from \citet{puebla2022can}, from which they conclude that DNNs can learn generalizable sameness visual relations. However, they only tested this model on nine of the 13 out-of-distribution datasets used by \citet{puebla2022can}. Notably, the four untested datasets were amongst the most difficult to generalize to in \citet{puebla2022can}'s simulations.

The third and fourth models are based on the Slot Attention architecture \citep[][see Fig. \ref{fig:models} top-right]{locatello2020object}. Through an iterative process, this module maps the feature map from a CNN to a set of slots representing individual objects and the background. It does this by calculating key-query-value attention weights between the vectors at different positions of the CNN feature map (keys and values), and the current slot vectors (queries). These attention weights are normalized over slots, which forces them to compete in order to \say{explain} the input. At each iteration, each slot is updated by a gated recurrent unit (GRU) layer, which takes as input the current slot vector and the attention vectors. The final slot vectors are combined to produce an image reconstruction, which is used to train the model through a reconstruction loss. Since Slot Attention was introduced as a way to enable abstract visual reasoning in DNNs, we included two models based based on this architecture.

The third model is a simple extension of the Slot Attention module that processes the final slots through a Relation Network \cite[RN;][see Fig. \ref{fig:models} top-right]{santoro2017simple}. This RN module takes each permutation of two slots and applies a single feed-forward network $g_{\theta}$ to obtain pairwise relation embeddings, which are then summed up and processed by another feed-forward network $g_{\phi}$ to obtain the final prediction. We used this simple extension (henceforth, SA-RN) to investigate the degree to which the the final slots of the Slot Attention module are sufficient to learn and generalize visual relations. 

The fourth model, Object-Centric Relational Abstraction\footnote{Since the models of \citet{adeli2023brain} and \citet{webb2023systematic} share the same acronym, we are going to use OCRAbs to refer to Webb et al.'s model and keep OCRA for Adeli et al.'s model since it was published first.} \citep[OCRAbs;][]{webb2023systematic}, is a specialized visual reasoning architecture that operates over the representations generated by a Slot Attention module (see Fig. \ref{fig:models} bottom-left). In this model processing starts with the extraction of feature ($z_k$) and position ($m_k$) embeddings by applying the attention maps of the slots to the feature maps from a CNN and its associated positional encoding. Then, OBCRAbs uses a relational operator to compute the pairwise relations between feature embeddings: $\phi(z_k, z_{k'}) = (z_k W_z \cdot z_{k'}W_z)W_r$. The dot product in this relational operator is meant to represent the relations between objects independently from their individual features. In fact, \citet{webb2023systematic} conjecture that this operator should improve generalization to relational samples with novel perceptual features. Finally, in order to represent higher-order relations, the pairwise relations $\phi(z_k, z_{k'})$ are summed up with linear projections of their corresponding objects' position embeddings and passed to a transformer that yields the model's output. \citet{webb2023systematic} showed that this model is capable of learning all 23 tasks of the SVRT as well as showing out-of-distribution generalization in a simplified version of the RMTS task (see section \ref{section:simulations} for details). From these results, \citet{webb2023systematic} conclude that OCRAbs is capable of learning relational representations that support systematic generalization of abstract rules.

The fifth model is Object-Centric Recurrent Attention \citep[OCRA;][see Fig. \ref{fig:models} middle-left]{adeli2023brain}. This model is a encoder–decoder architecture that combines a recurrent glimpse-based attention mechanism with capsule-based object-centric representations. At each time step of the encoder processing, a read glimpse is processed by a small CNN that feeds into a encoder RNN. A two-layer capsule network reads the encoder activations in the first layer and routs them through agreement to the class capsules in the second layer. The magnitude of these capsules corresponds to the evidence for a class in the current glimpse. In the decoder part of the model, the most active class capsule is routed to the decoder RNN, which generates the read \say{where} and \say{what} parameters that determine the area selected by the read attention mechanism. The decoder also generates \say{where} and \say{what} parameters for an analogous write mechanism that iteratively reconstructs the canvas. \citet{adeli2023brain} showed that OCRA outperformed a ResNet18 model in same-different generalization using a testing protocol from \citet{puebla2022can}, where the models are trained in nine different datasets and tested on four withheld datasets. From these results, \citet{adeli2023brain} concluded that OCRA's glimpse-based attention mechanism is important for achieving visual reasoning in DNNs. Note, however, that OCRA's same-different generalization was still limited.

The sixth model was Guided Attention Model for (visual) Reasoning \citep[GAMR;][see Fig. \ref{fig:models} middle-left]{vaishnav2023gamr}. In this model a recurrent controller shifts attention to relevant image locations and stores attended object representations in a memory bank. At the same time, the controller interacts with the memory bank to guide attention to the next relevant location for the task at hand. GAMR consist of three modules. The encoder module consist of a CNN that generates a low dimensional representation of the image $z_{img}$. The controller module consist of two blocks. The first one is a LSTM that generates an internal query to guide attention towards task relevant features. The second one is a guided attention block that extracts relevant visual information from the image conditional on the internal query at each time step $t$. This block generates a context vector $z_t$ that is stored in the memory bank. Finally, the reasoning module is similar to the aforementioned RN \citep{santoro2017simple}. It consists of a two-layer MLP $r_\theta$ that produces an aggregated relational embedding. This relational embedding is concatenated with the output of the LSTM controller at the last time step of processing and passed through a decision layer $f_\phi$ to predict the output for the task at hand. Among other results, \citet{vaishnav2023gamr} showed that GAMR was able to learn all 23 tasks of SVRT, achieving the best overall performance of all models tested. Furthermore, GAMR was able to generalize without training its learning to other related tasks. For example, when trained on SVRT task 1 (the standard SD task), GAMR achieved $84.91\%$ test accuracy on SVRT task 22, which consisted of deciding whether in a canvas of four shapes there were two pairs of identical shapes or not. From these results, \citet{vaishnav2023gamr} concluded that GAMR was capable of performing abstract visual reasoning.

\section{Simulations}\label{section:simulations}

\subsection{Simulation 1: Match-to-sample}

\begin{figure}[t]
    \centering
    \includegraphics[width=\linewidth]{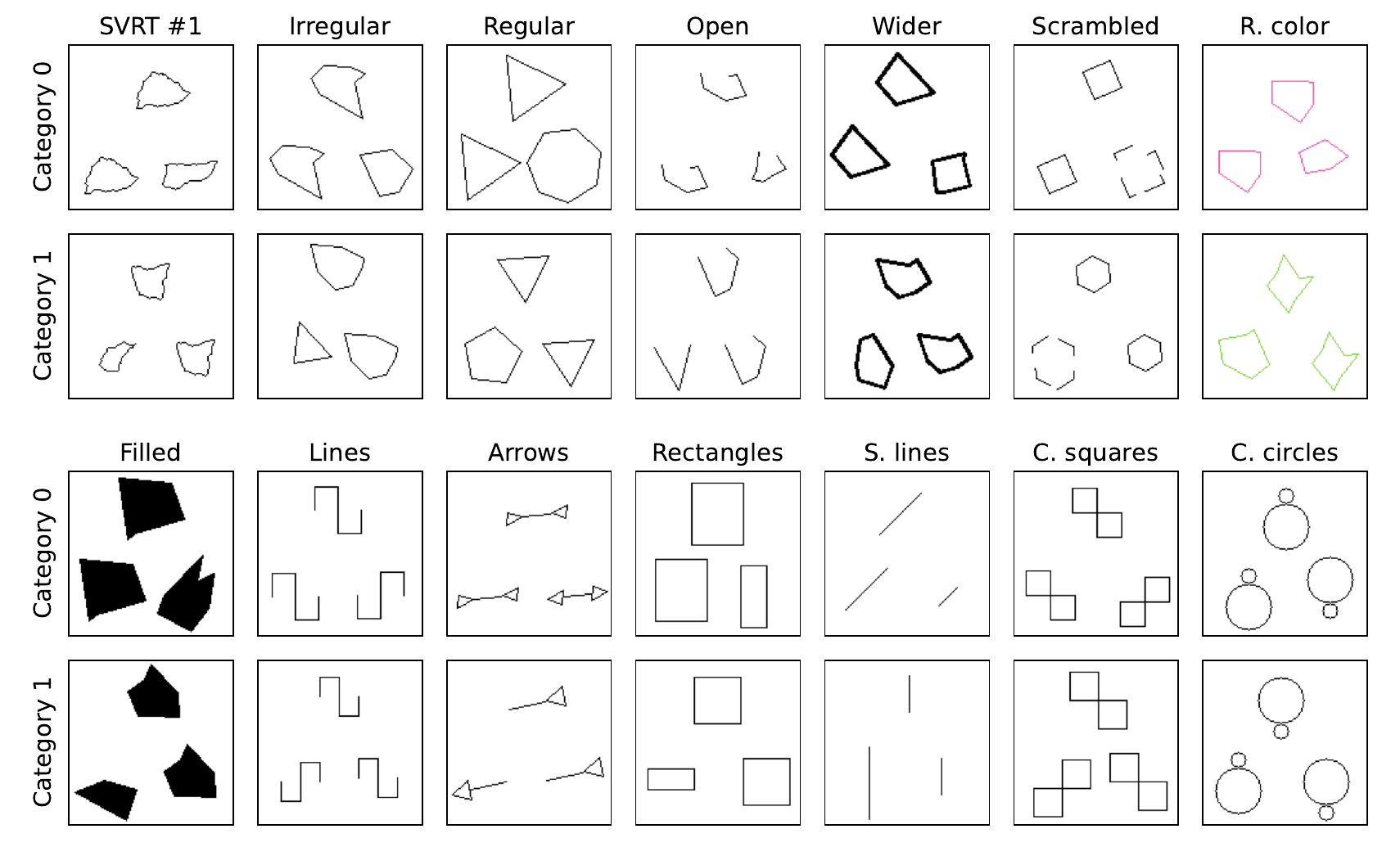}
    \caption{Positive and negative MTS examples per dataset.}
    \label{fig:examplesMTS}
\end{figure}

\subsubsection{Data}

As described earlier, in the MTS task there are three shapes, one at the top-center (base) and two at each side of the bottom part of the canvas (candidates), and the goal is to identify which candidate shape matches the base. Following the testing protocol of \citet{puebla2022can}, we generated an \emph{Original} dataset of $128\times128$ images using shapes taken from the SVRT \citep{fleuret2011comparing} and 13 out-of-distribution datasets. The Original dataset consisted of splits of $28000$, $5600$ and $11200$ images for training, validation and testing respectively. On all MTS datasets, half of the images corresponded to cases where the match was to the left. In order to enable translation there were three subareas of $64\times64$ pixels (one at the top-center, one at the bottom-left and one at the bottom-right) where the shapes were placed randomly with the constrain that they did not touch the canvas limits. 

The 13 out-of-distribution datasets were created following the same abstract rule, but were implemented using shapes with different perceptual features than the Original dataset (see Figure \ref{fig:examplesMTS}). For example, in the Filled dataset the shapes were filled with black instead of being contours. For all the out-of-distribution datasets we also generated training, validation and testing splits of the same size as the Original dataset, but only used the test splits in Simulations 1-4. (We used the train and validation splits of 9 these out-of-distribution datasets in Simulation 5, where we explored the effects of a richer training environment on generalization.)

Next we describe each out-of-distribution dataset main features briefly (all the code necessary to generate these datasets can be found at the article's repository). In the \emph{Irregular} dataset each shape was a irregular polygon generated by sampling a series of 3 to 8 points around a randomly chosen center. After this, we added uniformly distributed random noise to each point and connected all of them with straight lines. In the \emph{Regular} dataset each shape was a regular polygon. These polygons were generated in the same way as the Irregular dataset except that we did not add random noise to the polygon points. The \emph{Open} dataset was generated in the same way as the Irregular dataset except that the first and last vertices of each shape were not connected. The \emph{Wider} dataset was generated in the same way as the Irregular dataset except that the line width was set to two pixels instead of one. The \emph{Scrambled} dataset was was generated in the same way as the Regular dataset except that one the of the shapes was scrambled by dividing it into sections and displacing them randomly around the center. The \emph{Random Color} dataset was generated in the same way as the Irregular dataset except that for each image the line color was chosen randomly. The \emph{Filled} dataset was generated in the same way as the Irregular dataset except that the shapes were filled with black. In the \emph{Lines} dataset each object corresponded to a line created by joining two open squares, one with the opening pointing downward and the other with the opening pointing upward, at the end of the opposite left/right sides. In the \emph{Arrows} dataset the objects were arrows consisting of one or two triangular head(s) and a line, the head(s) and the line were connected. In the \emph{Rectangles} dataset each shape was a rectangle. In a given image these rectangles could vary on height or with but not both at the same time. In the \emph{Straight Lines} dataset each shape was a horizontal straight line with a tilt of 0\degree, 45\degree, 90\degree or 135\degree. In a given image all lines varied on length but not tilt. In the \emph{Connected Squares} dataset each shape was a pair of connected squares. The shapes varied at the corner (left or right) at which both squares were connected. In the \emph{Connected Circles} dataset each shape was a pair of horizontally connected circles. One the circles was bigger than the other. The shapes varied on whether the small circle was on top of the big one or vice versa.

\subsubsection{Training}

We trained all models on the train split of the Original dataset and used its validation split for hyperparameter optimization. No data from the out-of-distribution datasets was used for training or validation. As our focus was on out-of-distribution generalization rather than training efficiency, we trained our models for a given maximum number of epochs (specific to each model) with an early stopping criterion of $99\%$ accuracy on the validation split of the Original dataset. We trained each model using 10 different random seeds and report the mean test accuracy on the Original and all out-of-distribution datasets. The ResNet-50 model was pretrained on ImageNet, CLIP-ViT was pretrained in a proprietary text-image dataset, and the Slot Attention module of OCRAbs was pretrained in all 23 tasks of SVRT (but as in \citealt*{webb2023systematic} was not modified during training). All other models were trained from scratch. Table \ref{table:trainingMTS} summarizes the training hyperparameters per model. All training details for each model are available in the article's repository.

\begin{figure*}[!ht]
    \centering
    \includegraphics[width=\linewidth]{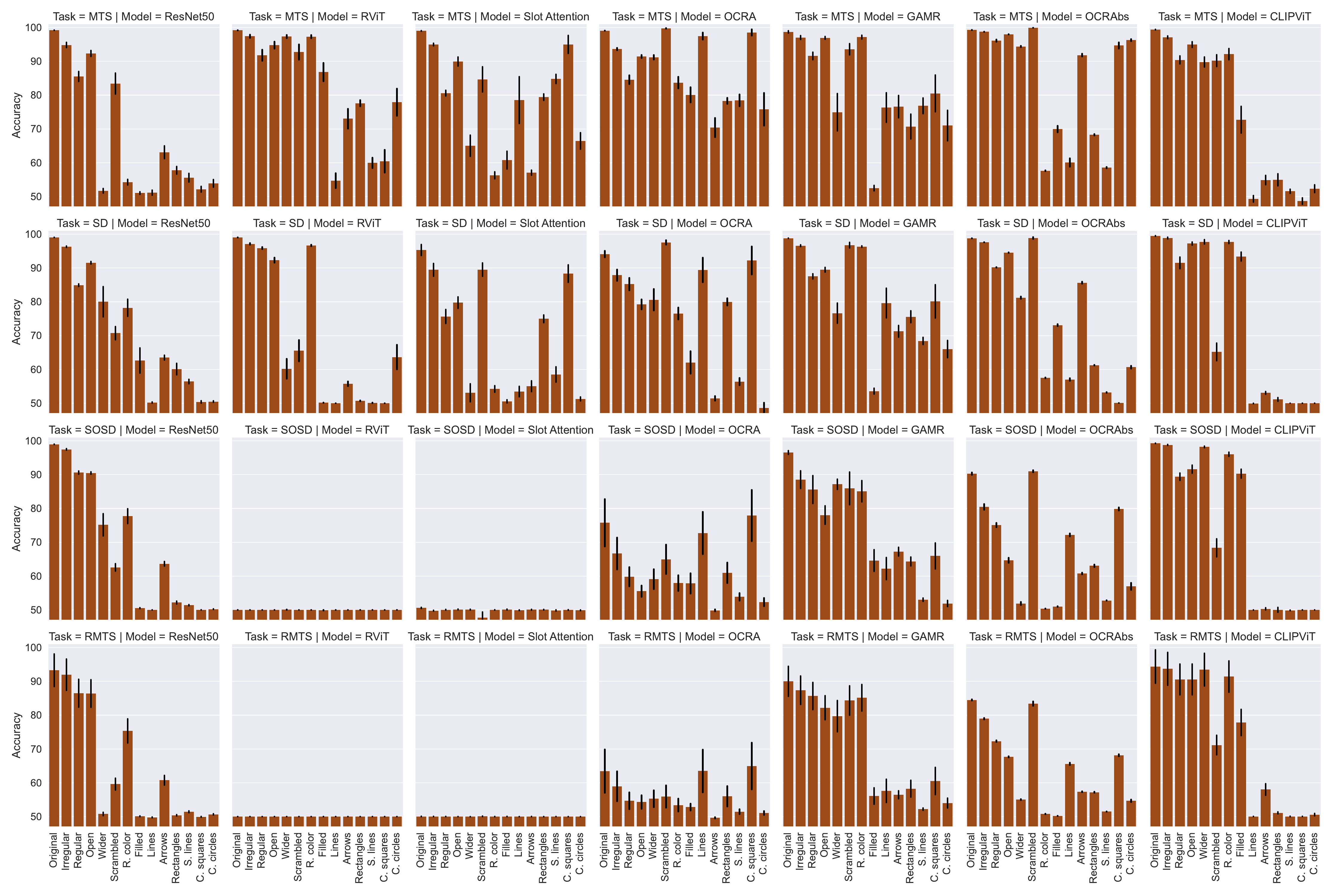}
    \caption{Accuracy by task, model and dataset. Error bars are standard errors of the mean.}
    \label{fig:results1-4}
\end{figure*}

\begin{table}[width=\linewidth,cols=4,pos=t]
\caption{Main MTS training hyperparameters per model.}
\label{table:trainingMTS}
\begin{tabular*}{\tblwidth}{@{} LLLL@{} }
    \toprule
    Model & Max. epochs & Init. lr & Opt. algorithm\\
    \midrule
    ResNet-50 & $100$ & $5e-6$ & Adam\\ 
    RViT & $100$ & $1e-4$ & Adam\\
    SA-RN & $300$ & $4e-4$ & Adam\\
    OCRA & $100$ & $7e-4$ & Adam\\
    GAMR & $100$ & $1e-5$ & Adam\\
    OCRAbs & $300$ & $4e-4$ & Adam\\
    CLIP-ViT & $100$ & $1e-6$ & AdamW\\
    \bottomrule
\end{tabular*}
\end{table}

\subsubsection{Results and discussion}

Our main results are presented in Figure \ref{fig:results1-4} first row. As can be seen, the baseline ResNet-50 model was able to learn the Original MTS dataset, achieving almost perfect in-distribution test accuracy ($99.2\%$). However, its average out-of-distribution performance across datasets was significantly worst ($65.1\%$), achieving a high ($>90\%$) accuracy only in the Irregular and Open datasets. In contrast, all object-centric models also showed an almost perfect in-distribution test performance (around $99\%$), but obtained comparatively higher out-of-distribution generalization performance (OCRA: $86.4\%$; OCRAbs: $83.4$, RViT: $81.7\%$; SA-RN: $81.7\%$; GAMR: $81.2\%$; CLIP-ViT: $72.2\%$). As can be appreciated in Figure \ref{fig:results1-4}, however, for each model there are important differences across datasets. For example, OCRA, the best performing model, obtained a high accuracy on six out of the 13 out-of-distribution datasets. Another notable case is CLIP-ViT, the worst performing model, which showed a marked difference in performance between the first six out-of-distribution datasets and the last seven.

\begin{landscape}
\begin{figure}[p]
    \centering
    \includegraphics[width=\linewidth]{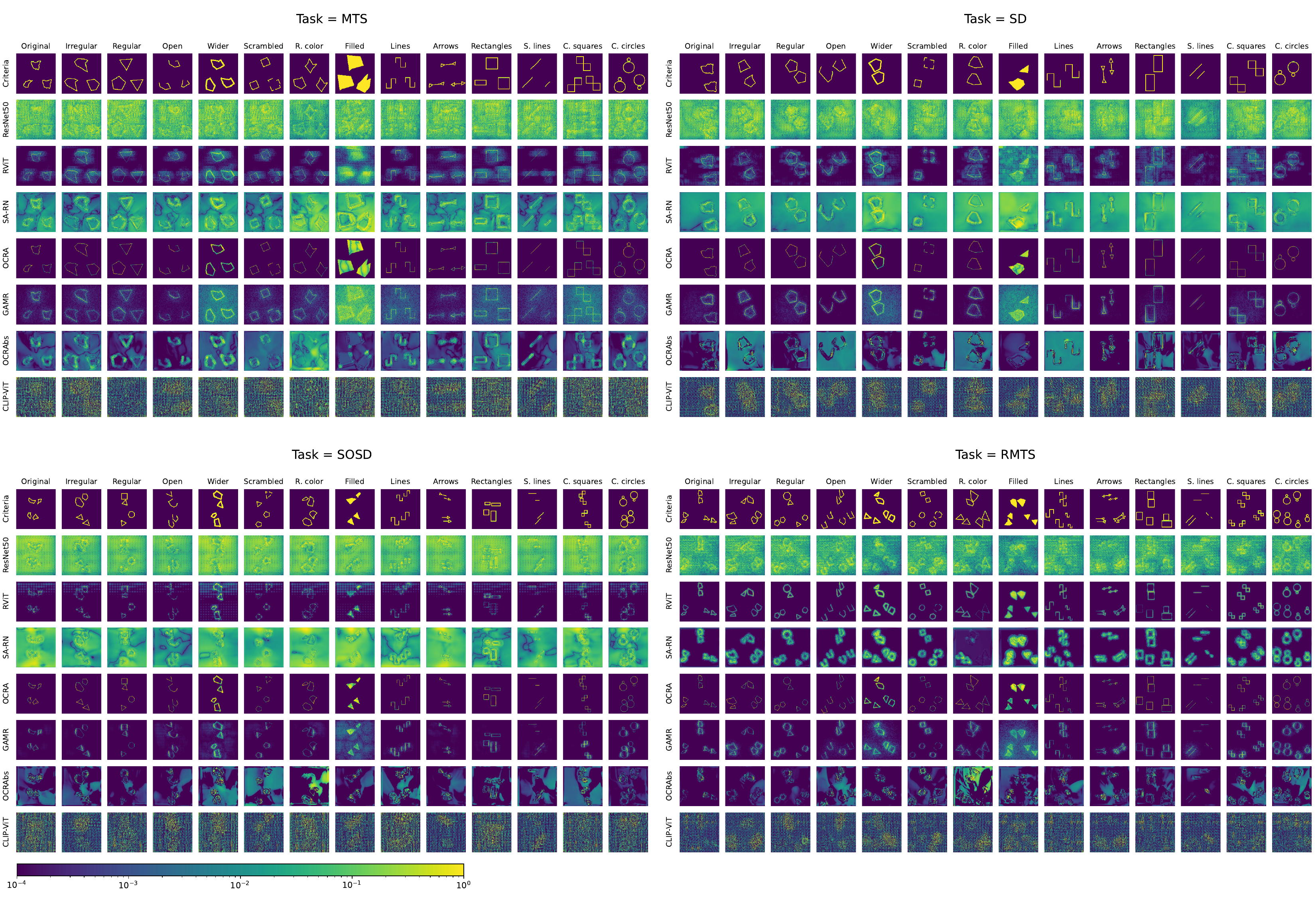}
    \caption{Model attributions by task, model and condition. For each task the first row shows the attributions of a simulated model with perfect object segregation. Model attributions were calculated through the integrated gradients method and a logarithmic scale clipped at $1e-4$ was used for the visualization.}
    \label{fig:attr}
\end{figure}
\end{landscape}

In order to evaluate the degree to which our models based their classification decisions on the images' objects instead of the whole canvas, we generated model attribution visualizations through the integrated gradients method \citep{sundararajan2017axiomatic}. This method approximates the integral of the gradients of the output of the model for the predicted class (i.e., the most active output unit) with respect to the input image features along the path from a baseline (in this case a black image) to the input image. Figure \ref{fig:attr} top-left panel shows the attribution visualization for a sample images from each dataset of the MTS task. To generate this visualizations we plotted the absolute value of the estimated integral of the gradients using a logarithmic scale. As can be seen, the baseline ResNet-50 model tended to consider the whole canvas when classifying the images, although some focus on the objects was appreciable. In contrast, RViT, SA-RN, OCRA, GAMR and OCRAbs, tended to focus more explicitly on the images' objects, which is consistent with previous research showing that the object segmentation performance of object-centric models is generally robust to local distribution shifts \citep{dittadi2022generalization}. In this task, CLIP-ViT based its classifications in the whole canvas but with some appreciable focus on the objects for some images.  

To sum up, our results show that, on the MTS tasks, object-centric models generalize better their learning to out-of-distribution samples than a standard CNN baseline. Although the overall out-of-distribution generalization performance of these models is similar, there are marked differences across datasets and models. Furthermore, with the exception of CLIP-ViT, we found that object centric models tended to base their classifications on the images' objects while ignoring the background. This supports the hypothesis that object-centric representations lead to better generalization in visual reasoning tasks. However, even the best performing object-centric model showed a significant drop in performance in more than half of out-of-distribution datasets, which suggests that object-centric representations are not a sufficient condition to achieve visual relational reasoning.

\subsection{Simulation 2: Same-different}

\subsubsection{Data}
\begin{figure}[!ht]
    \includegraphics[width=\linewidth]{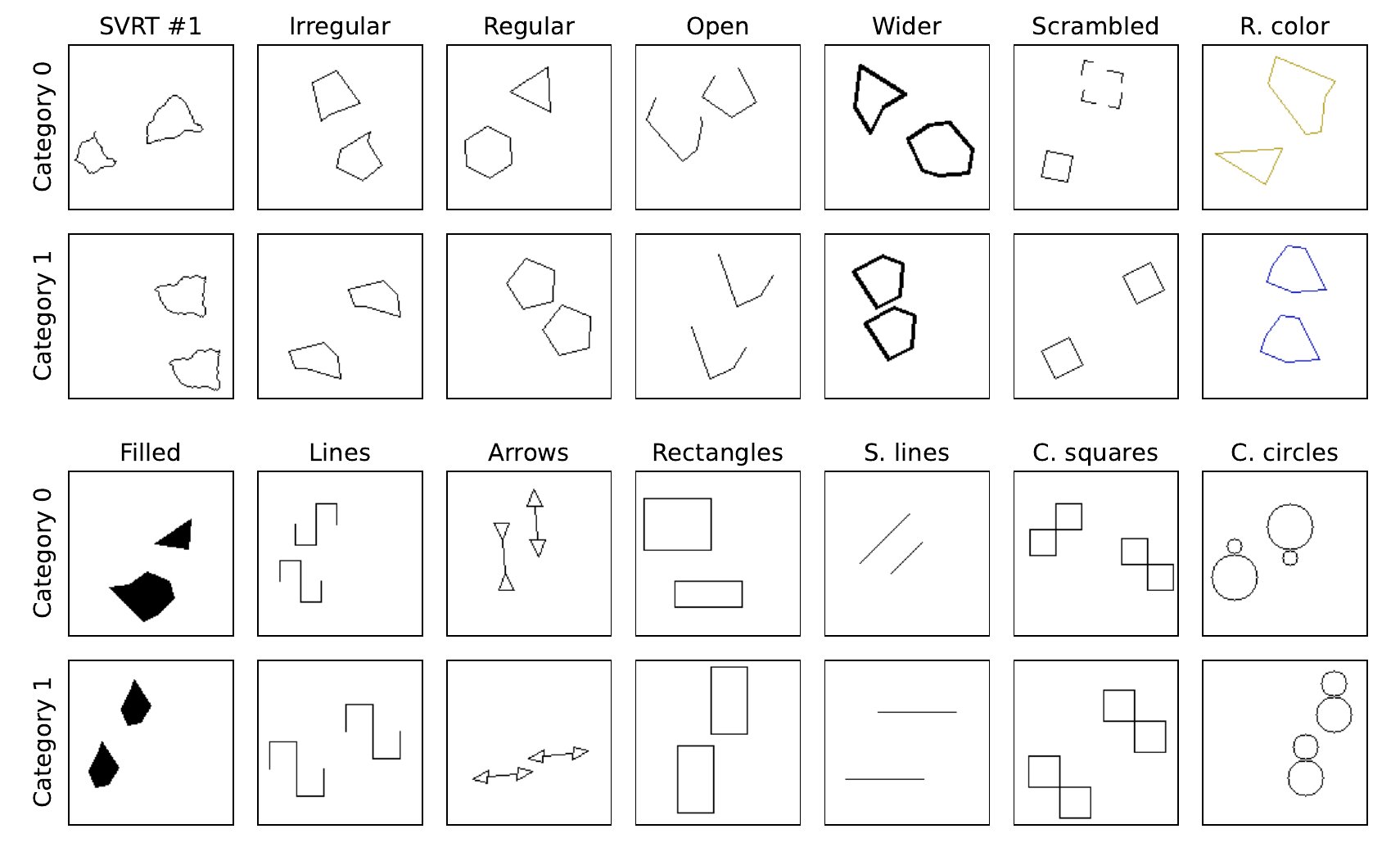}
    \caption{Same-different examples per condition.}
    \label{fig:examplesSD}
\end{figure}

As described earlier, the SD task involves classifying images of pairs of objects as \say{same} (labeled as $0$) or \say{different} (labeled as $1$). As in Simulation 1, we generated an Original dataset with shapes taken from the SVRT \citep{fleuret2011comparing} and 13 out-of-distribution datasets. Also as in Simulation 1, the out-of-distribution datasets were created following the same abstract rule as the Original dataset, but were implemented using shapes with different perceptual features than the Original dataset (see Figure \ref{fig:examplesSD}). 

All SD datasets consisted of splits of $28000$, $5600$ and $11200$ images for training, validation and testing, but we only used the test splits of the 13 out-of-distribution datasets for testing. On all SD datasets, half of the images corresponded to cases where the objects were the same (label 0).

\subsubsection{Training}
Training proceeded in the same way as in Simulation 1. Table \ref{table:trainingSD} presents the main training hyperparameters per model. All training details for each model are available in the article's repository.

\begin{table}[width=\linewidth,cols=4,pos=ht]
\caption{Main SD training hyperparameters per model.}
\label{table:trainingSD}
\begin{tabular*}{\tblwidth}{@{} LLLL@{} }
    \toprule
    Model & Max. epochs & Init. lr & Opt. algorithm\\
    \midrule
    ResNet-50 & $100$ & $1e-4$ & Adam\\ 
    RViT & $200$ & $1e-4$ & Adam\\
    SA-RN & $400$ & $4e-4$ & Adam\\
    OCRA & $100$ & $1e-3$ & Adam\\
    GAMR & $100$ & $1e-5$ & Adam\\
    OCRAbs & $300$ & $4e-4$ & Adam\\
    CLIP-ViT & $100$ & $1e-5$ & AdamW\\
    \bottomrule
\end{tabular*}
\end{table}

\subsubsection{Results and discussion}

Our main results are presented in Figure \ref{fig:results1-4} second row. Similarly to Simulation 1, the baseline ResNet-50 model was able to learn the Original SD dataset, achieving almost perfect in-distribution test accuracy ($99.0\%$). However, its average out-of-distribution performance across datasets was significantly worst ($68.9\%$), achieving a high ($>90\%$) accuracy only in the Irregular and Open datasets. In this task the object-centric models showed a more modest out-of-distribution generalization pattern compared to the MTS task, with four out of six models obtaining better generalization performance than the baseline (GAMR: $79.8\%$; OCRA: $75.9\%$; OCRAbs: $73.9$; CLIP-ViT: $72.7\%$; RViT: $67.5\%$; SA-RN: $67.2\%$). Again, there were important performance differences across datasets and models. For example, GAMR (the best performing model) obtained its worst accuracy on the Filled dataset. In comparison, CLIP-ViT (the worst performing model with better out-of-distribution generalization than the baseline), obtained a high test accuracy on this dataset.

As in Simulation 1, we obtained model attribution visualizations through the integrated gradients method (Figure \ref{fig:attr} top-right panel). As can be seen, the baseline ResNet-50 model tended to consider the whole canvas when classifying the images (again some focus on the objects was appreciable). In contrast, RViT, SA-RN, OCRA, GAMR and OCRAbs, tended to focus more explicitly on the images' objects (note, however, that SA-RN showed considerable more focus on the canvas than the rest of these models). Finally, we found a more clear focus on the images' objects for the CLIP-ViT model in comparison to the MTS task, although these visualizations are more diffused in comparison to the rest of the object-centric models.  

In summary, we found that four out of six object-centric models showed an overall better out-of-distribution generalization than the CNN baseline. However, this generalization advantage was less pronounced than in the MTS task. Furthermore, we found that same-different classification in object-centric models tended to based on the images' objects instead of the whole canvas. These results still lend support to the idea that of object-centric representations play an important role on visual reasoning. However, they suggest that this advantage depends critically on the task setting.

\subsubsection{Other models}
Previous research by \citet{webb2021emergent}, had claimed that the Emergent Symbol Binding Network (ESBN) was capable of learning abstract rules from images. Using a custom dataset of 100 simple icons, they showed that ESBN could generalize several visual reasoning tasks, including the SD task, to unseen out-of-distribution samples. We tried to train this model in our Original SD dataset but quickly realised that it was not capable of learning it. In Appendix \ref{section:ESBN} we show that this model does not learn abstract representations of rules as originally claimed, but instead uses a shortcut to leads to perfect generalization in their custom dataset. 

\subsection{Simulation 3: Second order same-different}

While Simulations 1 and 2 focused on sameness tasks that entails a judgment on the relation between pairs of objects, Simulations 3 and 4 dealt with the SOSD and RMTS tasks, that requires comparisons between relations. We start with the SOSD task, which is the task with less objects and therefore potentially the simplest of the two.

\begin{figure}[t]
    \centering
    \includegraphics[width=\linewidth]{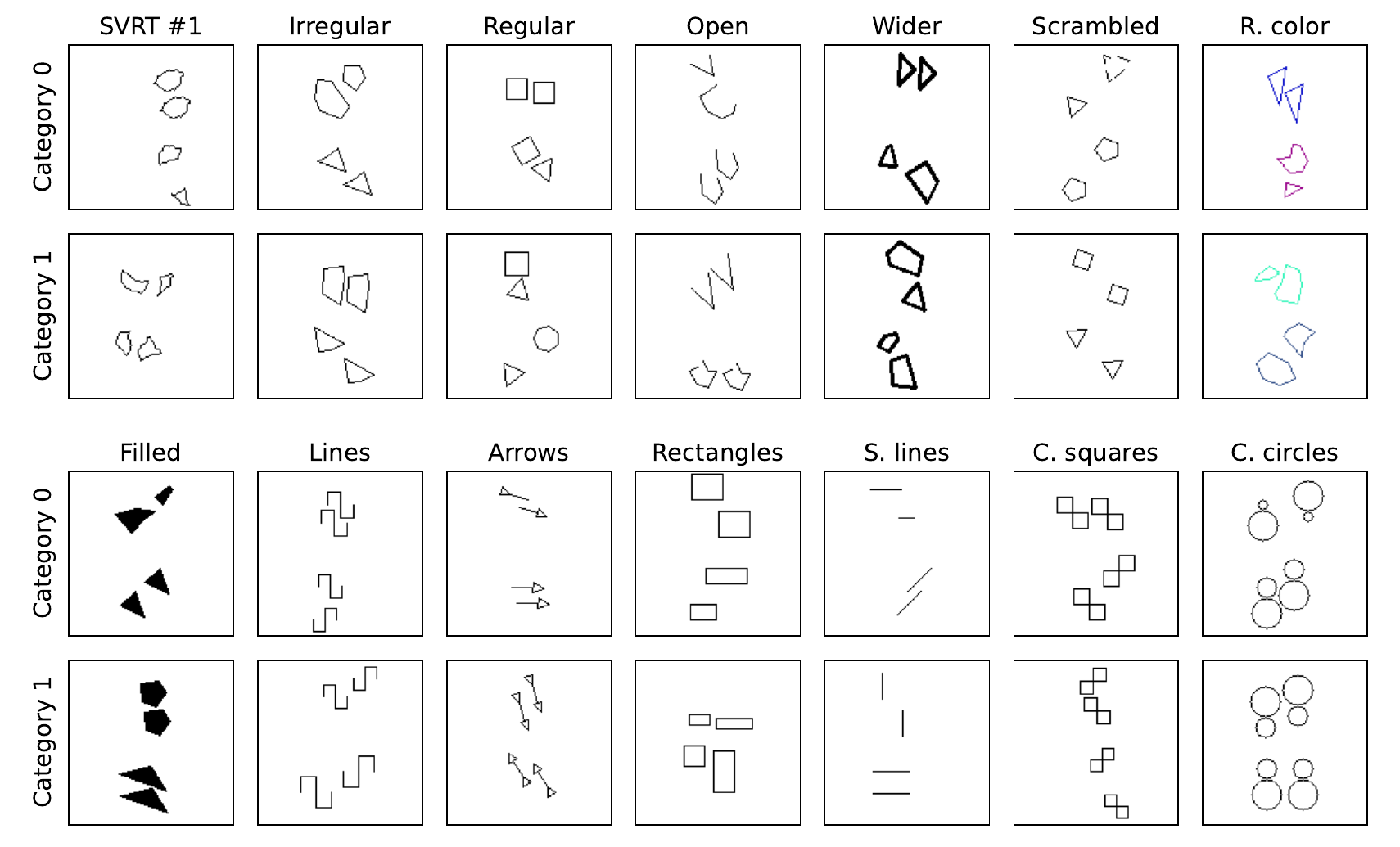}
    \caption{Second order same-different examples per condition.}
    \label{fig:examplesSOSD}
\end{figure}

\subsubsection{Data}

As described earlier, the SOSD task involves deciding whether two pairs of objects, one pair at the top-center and the other pair at the bottom-center, exemplify a single relation (i.e., \say{same-same} or \say{different-different}; labeled as 1) or two different ones (i.e., \say{same-different} or \say{different-same}; labeled as 0). As in Simulations 1 and 2, we generated an Original dataset with shapes taken from the SVRT \citep{fleuret2011comparing} and 13 out-of-distribution datasets. The 13 out-of-distribution datasets were created following the same abstract rule as the Original dataset, but were implemented using shapes with different perceptual features than the Original dataset. This perceptual features were the same as in Simulations 1 and 2 (see Figure \ref{fig:examplesSOSD}). All SOSD datasets consisted of splits of $98000$, $14000$ and $28000$ images for training, validation and testing, but we only used the test splits of the 13 out-of-distribution datasets for testing. On all SOSD datasets, half of the images corresponded to cases where the objects exemplified two different relations (label 0).

\subsubsection{Training}
Training proceeded in the same way as in Simulations 1 and 2. Table \ref{table:trainingSOSD} presents the main training hyperparameters per model. All training details for each model are available in the article's repository.

\begin{table}[width=\linewidth,cols=4,pos=ht]
\caption{Main SOSD training hyperparameters per model.}
\label{table:trainingSOSD}
\begin{tabular*}{\tblwidth}{@{} LLLL@{} }
    \toprule
    Model & Max. epochs & Init. lr & Opt. algorithm\\
    \midrule
    ResNet-50 & $100$ & $5e-5$ & Adam\\ 
    RViT & $100$ & $1e-4$ & Adam\\
    SA-RN & $400$ & $1e-4$ & Adam\\
    OCRA & $200$ & $7e-4$ & Adam\\
    GAMR & $150$ & $1e-5$ & Adam\\
    OCRAbs & $100$ & $4e-6$ & Adam\\
    CLIP-ViT & $100$ & $1e-6$ & AdamW\\
    \bottomrule
\end{tabular*}
\end{table}

\subsubsection{Results and discussion}

Our main results are presented in Figure \ref{fig:results1-4} third row. Similarly to Simulations 1 and 2, the baseline ResNet-50 model was able to learn the Original SOSD dataset, achieving almost perfect in-distribution test accuracy ($99.0\%$). Also as in Simulations 1 and 2, its average out-of-distribution performance across datasets was significantly worse ($66.3\%$), achieving a high ($>90\%$) accuracy only in the Irregular and Open datasets. In contrast with the MTS and SD tasks, RViT and SA-RN were not able to learn the SOSD task (both in-distribution test accuracies around $50\%$) and OCRA did not achieve a high in-distribution test accuracy ($75.8\%$). Furthermore, the object-centric models showed an even more modest out-of-distribution generalization pattern than in the SD task, with two out of six models obtaining better overall generalization performance than the baseline (GAMR: $72.3\%$; CLIP-ViT: $71.8\%$; OCRAbs: $65.4$; OCRA: $60.7\%$; SA-RN: $50.0\%$; RViT: $49.8\%$). Again, there were important differences across datasets and models.

\begin{figure}[!t]
    \includegraphics[width=\linewidth]{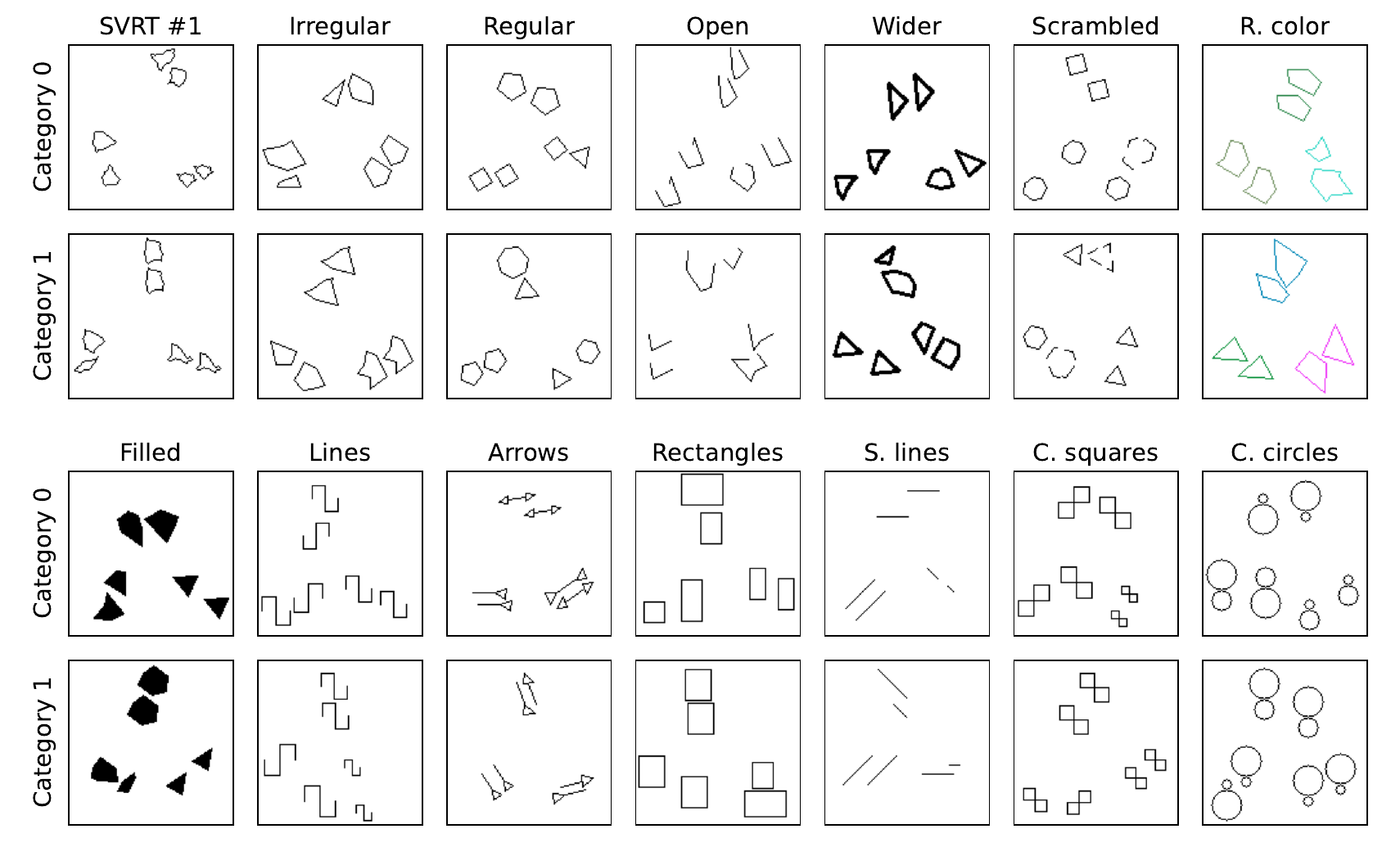}
    \caption{Relational match-to-sample examples per condition.}
    \label{fig:examplesRMTS}
\end{figure}

As in Simulations 1 and 2, we obtained model attribution visualizations through the integrated gradients method (Figure \ref{fig:attr} bottom-left panel). As can be seen, the baseline ResNet-50 model tended to consider the whole canvas when classifying the images (again some focus on the objects was appreciable). In contrast, RViT, SA-RN, OCRA, GAMR and OCRAbs, tended to focus more explicitly on the images' objects (again SA-RN showed a higher focus on the background than the rest of the object-centric models). Finally, for the CLIP-ViT model we found a similar level of focus on the images's objects to the SD task.  

To sum up, we found that the SOSD was in general harder for the object centric models, with two models not being able to learn the task and with two of the remaining models showed an overall better out-of-distribution generalization than the CNN baseline. Furthermore, the generalization advantage of the two successful models (GAMR and CLIP-ViT) was significantly less pronounced than in the SD and MTS task. In addition, we found that, with the exception of SA-RN and CLIP-ViT, the object-centric models tended to base their classifications on the images' objects instead of the whole canvas. These results further support the hypothesis that the task setting plays a critical role on the benefits that object-centric representations can exert of the relational visual reasoning capabilities of DNNs.

\subsection{Simulation 4: Relational match-to-sample}
\subsubsection{Data}
As described earlier, in the RMTS task there is a base pair of objects exemplifying a relation (\say{same} or \say{different}) at the top-center of the canvas and two candidate pairs (bottom-left and bottom-right) and the goal is to identify which candidate matches the relation of the base pair. The image is labeled as $0$ if the match is to the bottom-left and as $1$ if the the match is to the bottom-right. As in Simulations 1-3, we generated an Original dataset with shapes taken from the SVRT \citep{fleuret2011comparing} and 13 out-of-distribution datasets. The 13 out-of-distribution datasets were created following the same abstract rule as the Original dataset, but were implemented using shapes with different perceptual features than the Original dataset. This perceptual features were the same as in Simulations 1-3 (see Figure \ref{fig:examplesRMTS}). All RMTS datasets consisted of splits of $196000$, $28000$ and $56000$ images for training, validation and testing, but we only used the test splits of the 13 out-of-distribution datasets for testing. On all RMTS datasets, half of the images corresponded to cases where the match was to the bottom-left.

\subsubsection{Training}

Training proceeded in the same way as in Simulations 1-3. Table \ref{table:trainingSOSD} presents the main training hyperparameters per model. All training details for each model are available in the article's repository.

\begin{table}[width=\linewidth,cols=4,pos=ht]
\caption{Main RMTS training hyperparameters per model.}
\label{table:trainingRMTS}
\begin{tabular*}{\tblwidth}{@{} LLLL@{} }
    \toprule
    Model & Max. epochs & Init. lr & Opt. algorithm\\
    \midrule
    ResNet-50 & $200$ & $5e-5$ & Adam\\ 
    RViT & $100$ & $1e-4$ & Adam\\
    SA-RN & $400$ & $1e-4$ & Adam\\
    OCRA & $200$ & $7e-4$ & Adam\\
    GAMR & $200$ & $1e-5$ & Adam\\
    OCRAbs & $200$ & $1e-3$ & Adam\\
    CLIP-ViT & $100$ & $1e-6$ & AdamW\\
    \bottomrule
\end{tabular*}
\end{table}

\subsubsection{Results and discussion}

Our main results are presented in Figure \ref{fig:results1-4} fourth row. Similarly to Simulations 1-3, the baseline ResNet-50 model was able to learn the Original SOSD dataset, although is this achieving only a high level in-distribution test accuracy ($93.3\%$). The average out-of-distribution performance of this models was lower than in Simulations 1-3 ($62.6\%$), achieving a high ($>90\%$) accuracy only in the Irregular dataset. As in the SOSD task, RViT and SA-RN were not able to learn the RMTS task (both in-distribution test accuracies $50\%$) and OCRA did not achieve a high in-distribution test accuracy ($63.4\%$). Furthermore, the object-centric models showed an even more modest out-of-distribution generalization pattern than in the SOSD task, with two out of six models obtaining better overall generalization performance than the baseline (CLIP-ViT: $70.7\%$; GAMR: $69.2\%$;  OCRAbs: $62.5$; OCRA: $55.5\%$; SA-RN: $50.0\%$; RViT: $50.0\%$). Again, there were important performance differences across datasets and models.

As in Simulations 1-3, we obtained model attribution visualizations through the integrated gradients method (Figure \ref{fig:attr} bottom-right panel). Consistently with the previous simulations, the baseline ResNet-50 model tended to consider the whole canvas when classifying the images (again some focus on the objects was appreciable). In contrast, RViT, SA-RN, OCRA, GAMR and OCRAbs, tended to focus more explicitly on the images' objects (in contrast with previous simulations SA-RN showed a more clear focus on the objects in comparison to the background). Finally, for the CLIP-ViT model we found a similar level of focus on the images' objects to the previous tasks.  

To sum up, we found that RMTS was the hardest of all our tasks in terms of both in-distribution and out-of-distribution performance. In this task four out of six object-centric models did not achieve a high in-distribution accuracy and the two remaining two model only achieve a slightly better out-of-distribution accuracy than the CNN baseline. As in previous simulations, we found that the object-centric models tended to base their classifications on the images' objects instead of the whole canvas. These results strongly suggest that object-centric representations are not a sufficient condition for DNNs to achieve relation-based generalization in visual reasoning tasks.

\subsection{Simulation 5: Rich training regime}

In our final simulation we explored the possibility that a rich training regime, composed of samples with varied perceptual features, would lead to improved out-of-distribution relational generalization for object-centric models. For this we followed the testing protocol of \citet[][Simulation 5]{puebla2022can}, where ResNet-based models were trained in 10 SD datasets (Original, Irregular, Regular, Open, Wider, Random Color, Filled, Lines and Arrows) and were tested in four out-of-distribution datasets (Rectangles, Straight lines, Connected Squares and Connected Circles). \citet{puebla2022can} found that a this training protocol did not result in out-of-distribution generalization for ResnNet-based models in the SD task. Here we applied this protocol in all our tasks to the ResNet50 baseline and the two best overall performers of Simulations 1-4, GAMR and RViT.

\begin{figure*}[t]
    \centering
    \includegraphics[width=\linewidth]{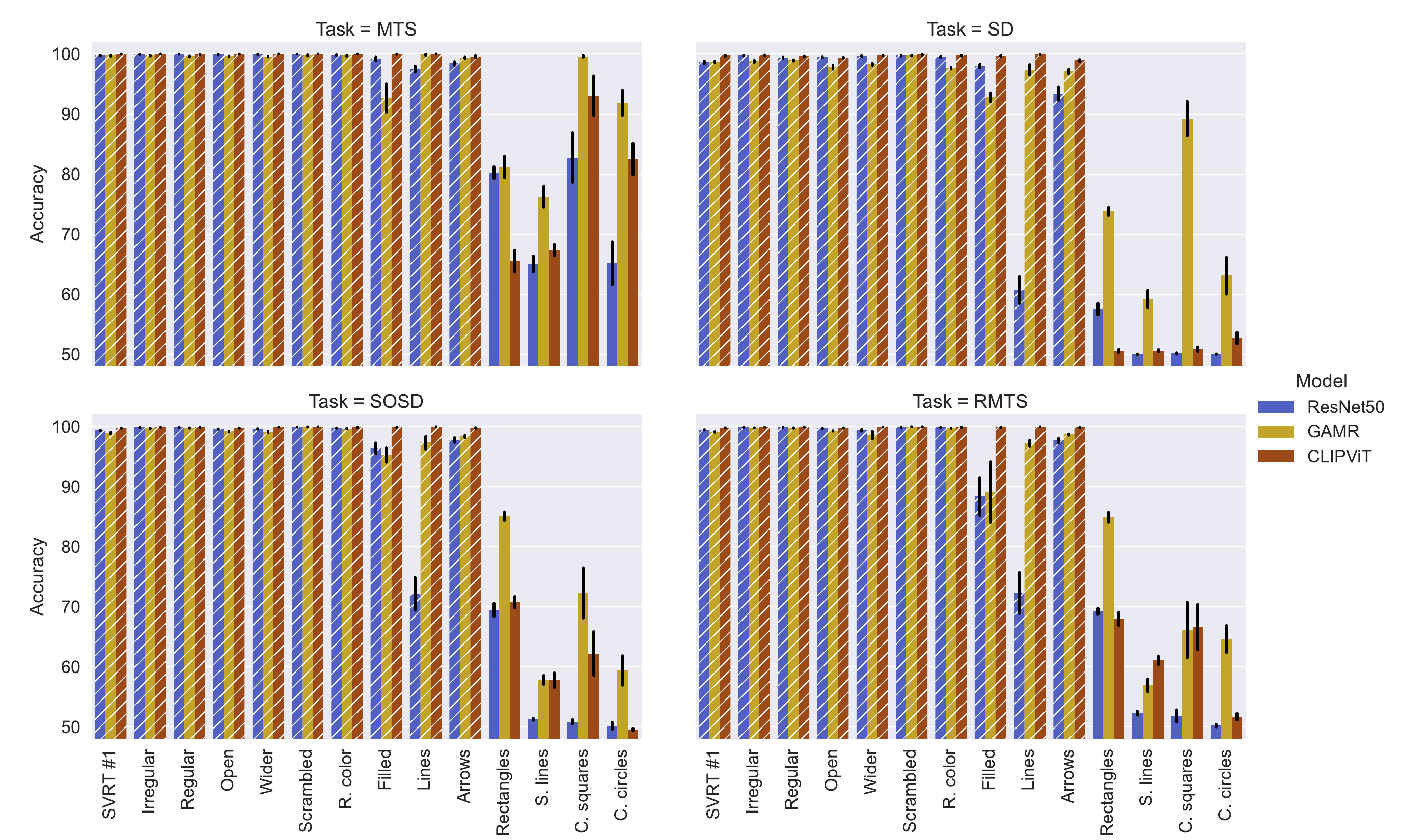}
    \caption{Accuracy by task, dataset and model in Simulation 5. Hatch marks indicate trained datasets. Error bars are standard errors of the mean.}
    \label{fig:MTLtest}
\end{figure*}

\subsubsection{Data}

For each task we generated a training dataset by concatenating the training splits of the Original, Irregular, Regular, Open, Wider, Random Color, Filled, Lines and Arrows datasets. We generated the validation datasets in the same way. We used the test splits of the Rectangles, Straight lines, Connected Squares and Connected Circles datasets as the out-of-distribution tests.

\subsubsection{Training}

As in \citet{puebla2022can}, during training on the composed datasets we randomly sampled batches from each individual dataset (i.e., each batch had data from a single dataset). Otherwise, training proceeded in the same way as in the previous simulations. We trained each model for a maximum of 20 epochs on the composed datasets. The rest of the hyperparameter for each model was the same as in the previous simulations.

\subsubsection{Results and discussion}

Our main results are presented in Figure \ref{fig:MTLtest}. As can be seen, all models achieved a high in-distribution test accuracy in most datasets, with the exception of the ResNet50 model in the Lines dataset of the SD, SOSD and RMTS task, and in the Filled dataset of the RMTS task, and the GAMR model in the Filled dataset of the RMTS task. As can be seen in Figure \ref{fig:MTLtrain}, all model instances exhibited a non-decreasing validation loss curve by epoch 20 or reached the early stop criterion of $99.0\%$ validation accuracy, which confirms that our models are fully trained. 

In terms of out-of-distribution generalization, we found a clear advantage for the MTS task, specially for GAMR and CLIP-ViT. In the rest of the tasks there was a clear drop in performance for the CLIP-ViT model. The best overall performer model was GAMR, which still exhibited a marked drop in performance in comparison to the in-distribution datasets, with the SD task showing worse generalization than MTS, and SOSD and RMTS showing the worst out-of-distribution generalization, similarly to Simulations 1-4.    

In summary, we found that a rich training regime did not lead to out-of-distribution generalization in visual reasoning tasks. These results replicate and extend the ones of \citet{puebla2022can}, who found that a rich training regime did not lead to out-of-distribution same-different discrimination of ResNet-based models. Furthermore, these results further support the hypothesis that different visual reasoning task impose different representational demands for DNNs, with a out-of-distribution generalization pattern for the best performing model, GAMR, similar to the one identified in comparative cognition research.

\section{General discussion}

Relational reasoning is fundamental to human thought. This has motivated a body of research into the capacity of DNNs to learn visual relations from image data. In particular, many articles have focused on the question whether DNNs can learn the concept of sameness \citep{adeli2023brain,baker2023configural,funke2021five,kim2018not,messina2021solving,messina2022recurrent,puebla2022can,ricci2021same,stabinger2021arguments,tartaglini2023deep,vaishnav2022understanding,webb2023systematic,webb2021emergent}. A key limitation of this work, however, is that most models are assessed with in-distribution test images, and in most cases, on a single task. In contrast, in this work, we focused in out-of-distribution generalization as a way to asses whether the representations learned by these models are actually relational. Furthermore, we used a range of sameness task derived from the comparative cognition literature. We focused on object-centric DNNs as there are theoretical and empirical reasons to suspect that these models would fare better at visual reasoning than standard CNN architectures and some of these models have been claimed to perform abstract visual reasoning. Our results show that even in our simpler tasks no single model could generalize its learning purely on the bases of the sameness relation.  Instead, generalization in these models depended on the perceptual features of the objects present in the images. Furthermore, out-of-distribution generalization performance depended critically on the task setting, with all models performing significantly worse in the SOSD and RMTS tasks in comparison to MTS and SD. For example, the RViT and SA-RN models were not even able to learn the SOSD and RMTS tasks despite of showing out-of-distribution generalization comparable to the rests of the models in the MTS and SD tasks. These results highlight the need to engage in severe testing before ascribing human-like cognitive abilities to DNNs \citep{bowers2023importance}.

Our results cast serious doubts on claims regarding CLIP-pretrained ViT models being able to learn \say{generalizable} sameness relations \citep{tartaglini2023deep}. Even on the standard SD task, CLIP-ViT was not able to generalize to a high level to seven out of our 13 out-of-distribution datasets. This pattern of generalization was fairly stable across tasks, with CLIP-ViT exhibiting slightly worse performance in the RMTS task than in the rest of the tasks. Interestingly, introducing a richer training regime only had a small effect on the MTS task, with this model achieving poor out-of-distribution generalization in all tasks. Another important result is that our model attribution visualizations showed a much more diffused focus on the images' objects than in the rest of the tested models. This suggests that CLIP-ViT might be operating differently than object-centric models, although further research is needed to draw conclusions in this regard.

The pattern of out-of-distribution of generalization obtained for the object centric models, as well as our model attribution visualizations, suggests that object-centric representations are not a sufficient condition to achieve relational generalization in the visual domain. Furthermore, our results suggests that current architectures are not capable of abstracting relational structure from object-centric representations of a scene. For example, OCRA, GAMR and OCRAbs have specialized modules designed to perform reasoning over perceptually segregated objects. However, in the SOSD and RMTS tasks, OCRA struggled to even learn the Original dataset, and GAMR and OCRAbs showed out-of-distribution generalization performance similar to the ResNet50 baseline. 

In the case of OCRAbs, our results directly contradict the claims of \citet{webb2023systematic}, who stated that its relational operator enables this model to form representations of \emph{abstract} relations. Importantly, similar claims have been made about ESBN, OCRAbs' predecessor \citep{webb2021emergent,webb2023relational}. As we show in appendix \ref{section:ESBN}, our results show that the claimed compositional generalization of ESBN can be better characterized as the model exploiting a shortcut in its training and test dataset.

More generally, our results speak to the debate about the binding problem in artificial neural networks. As has been noted in the cognitive science and AI literature \citep[e.g.,][]{greff2020binding,hummel2011getting}, processing relations between objects effectively in a neural architecture requires to form object-independent representations of relations, such that they can be applied freely to any object. It is this kind of representational flexibility that allows humans to generalize their learning in one task to perceptually dissimilar ones that share the same, or similar, abstract structure \citep{doumas2022theory}. As \citet{greff2020binding} point out, it is an open question how to integrate such binding mechanism between relations and objects with the deep learning framework. Historically, one known effective mechanism for binding in neural networks, neural synchrony, has been considered incompatible with gradient-based training, and therefore DNNs. However, recent developments on complex-valued activation functions might provide a foundation to integrate neural synchrony as a binding mechanism in DNNs \citep{lowe2022complex,stanic2023contrastive}.    

Repeatedly, several studies in vision science have established that the human visual system represents objects as configurations of relations between parts \cite[for a review see][]{hummel2013object}. In contrast, current object-centric representation learning methods do not make any attempt to capture this configural structure. Instead, their operation can be better described as perceptual segregation through clustering \citep{locatello2020object,mehrani5self}. Critically, there is substantial evidence that even the simple single-part abstract contour shapes used in the current study are represented by the human visual system as configurations of relations between segments of approximate constant curvature \citep{baker2018abstract,baker2021constantJEPG,baker2021constantPLOS}. It is likely that to capture the configural character of object representations, DNNs will need to integrate mechanisms to represent relations between parts of objects.

It is interesting to note that similarly strong claims regarding the relational reasoning capacities of DNNs in other domains have been falsified when tested more severely.  For instance, \citet{webb2023emergent} claimed that large language models are able to support abstract analogical reasoning.  However, \citet{lewis2024using} showed that good performance was lost when tested on out-of-distribution test stimuli.  In addition, various authors have argued that DNNs that learn disentangled representations are better able to support visual combinatorial generalisation \citep{higgins2018scan,watters2019spatial}.  However, once again, performance plummets when models are assessed on out-of-distribution test images \citep{montero2020role,montero2022lost}.  A central feature of human intelligence is that we can represent relations between entities in way that can be generalized across contexts.  Accordingly, it is necessary to assess models across tasks with out-of-distribution test stimuli before making claims regarding human-like relational reasoning in DNNs. 

To conclude, human visual reasoning is abstract in that it can generalize based on the relations between the objects in a scene instead of the objects themselves. Our results suggest that object-centric DNNs tend to perform well at segregating the objects in a scene, but their generalization performance is mainly driven by the surface features of the objects instead of their relations. It is likely that mechanisms to form independent object and relation representations and to compose them flexibly will be needed to achieve human-like visual reasoning.

\section{Declaration of competing interest}
The authors declare that they have no known competing financial interests or personal relationships that could have appeared to influence the work reported in this paper.

\section{Acknowledgments}

We would like to thank Hosein Adeli, Mohit Vaishnav, Tylor Webb, Alexa Tartaglini and Sheridan Feucht, who provided prompt access to the code base of their respective models.

The first author has received funding for this project from the National Center for Artificial Intelligence CENIA FB210017, Basal ANID. 

The second author has received funding for this project from the European Research Council (ERC) under the European Union’s Horizon 2020 research and innovation programme (grant agreement No 741134).

\appendix

\section{ESBN test}
\label{section:ESBN}

As explained in Simulation 2, \citet{webb2021emergent} claimed that the ESBN model is capable of almost perfect generalization of several visual reasoning tasks, including the standard SD task. ESBN is a recurrent architecture that processes images of individual objects sequentially. Its processing is characterized by two separate control and perception subsystems that interact through a external key-value memory. Critically, this interaction is driven by a dot-product-based comparison of the current image embedding, generated by the perception subsystem, and all other image embeddings stored in the external memory. As noted by \citet{kerg2022neural}, this dot product operation provides a string signal that can be exploited by a DNN, since any difference between two vectors will produce a pronounce drop in cosine similarity in comparison to two identical vectors. 

\citet{webb2023relational} argue that the interaction between control and perception in ESBN implements a \say{relational bottleneck} that encourages the model to identify relational patterns by abstracting away the details of specific examples seen during training. In contrast to these claims, \citet{vaishnav2023gamr} and \citet{kerg2022neural} have shown that ESBN is very brittle to non-relational manipulations such as translation, noise or background color changes. \citet{webb2023systematic} have taken these results as showing that ESBN \emph{assumes} object segregation (i.e., object-centric representations) and maintain that ESBN exhibits systematic generalization of relational rules. In this section we show evidence that the previously reported out-of-distribution generalization capabilities of ESBN are the result of a shortcut available on the icons dataset used to test it. Furthermore, we show that interpreting ESBN processing as assuming object-centric representations is incorrect.  

\begin{figure}[htp!]
    \centering
    \includegraphics[width=0.8\linewidth]{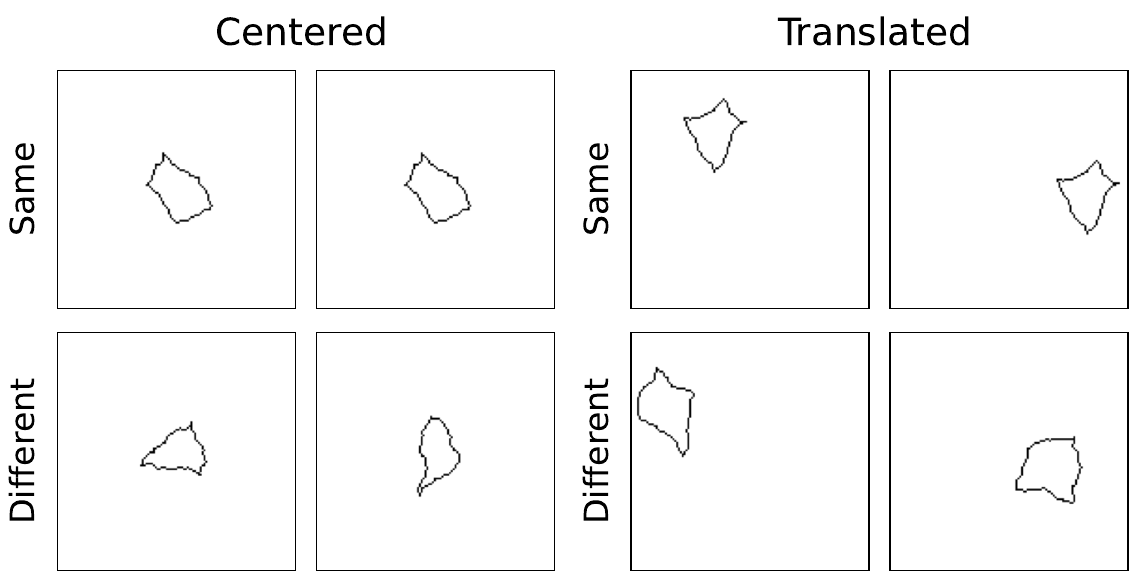}
    \caption{ESBN SD datasets.}
    \label{fig:ESBNstimuli}
\end{figure}

In our first simulation we generated two versions of the Original SD dataset, both with the objects presented in two separated images as needed for ESBN. As can be seen in Figure \ref{fig:ESBNstimuli}, in the \emph{centered} version all objects were centered in the canvas while in the \emph{translated} version the objects were placed at random positions (with the constraint that they didn't touch the canvas limits). Note that when the objects were the same in the centered version the images were \emph{exactly} the same. We trained 10 randomly initialized instances of ESBN in both datasets. We found that ESBN could learn the centered version in less than 20 epochs and could generalize perfectly to any of our out-of distribution datasets. However, the model was completely incapable of learning the translated version. This shows that ESBN can only handle pairs of images that are exactly the same or different at the pixel level. Note that this is a \emph{concrete} version of the SD task as opposed to the \emph{abstract} sameness relation that the SD task is meant to assess.

\begin{figure}[ht]
    \includegraphics[width=\linewidth]{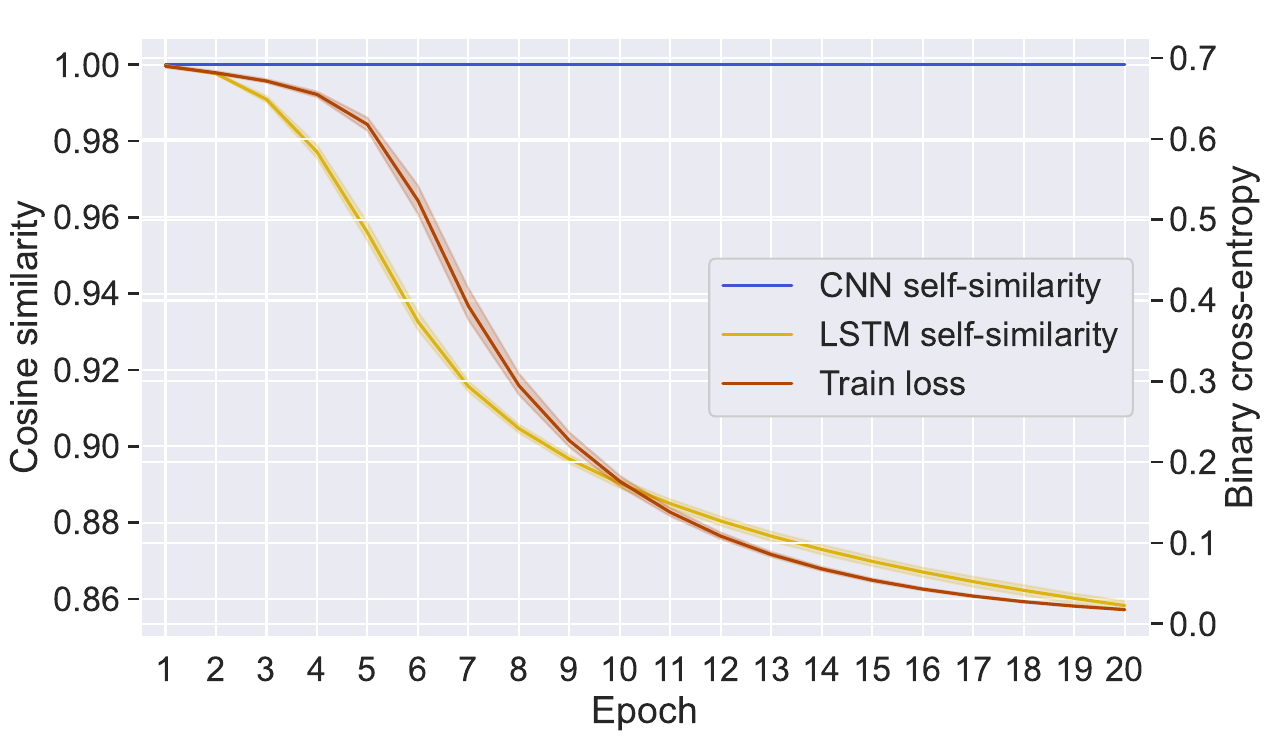}
    \caption{ESBN training analysis.}
    \label{fig:ESBNtrain}
\end{figure}

To gain further insights into this pattern of results, we recorded the initial weights of ESBN's CNN encoder and its LSTM controller. After each epoch of training, we calculated the cosine similarity of each component to its initial version before training. We plotted the resulting curves alongside the training loss in Figure \ref{fig:ESBNtrain}. As can be seen, training did not change ESBN's encoder, instead all learning was driven by changes in its LSTM controller. This clearly shows that ESBN exploited the strong signal given by the dot product of the encoded images. 

\begin{figure}[!ht]
    \includegraphics[width=\linewidth]{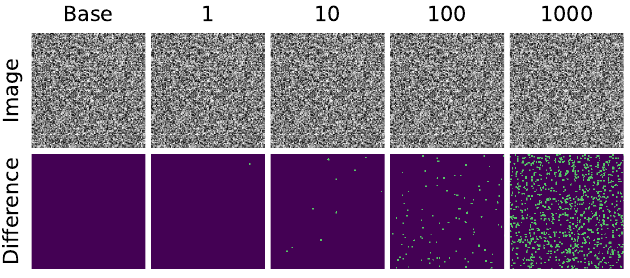}
    \caption{Random images stimuli.} 
    \label{fig:random_imgs}
\end{figure}

\begin{figure}[!ht]
    \includegraphics[width=\linewidth]{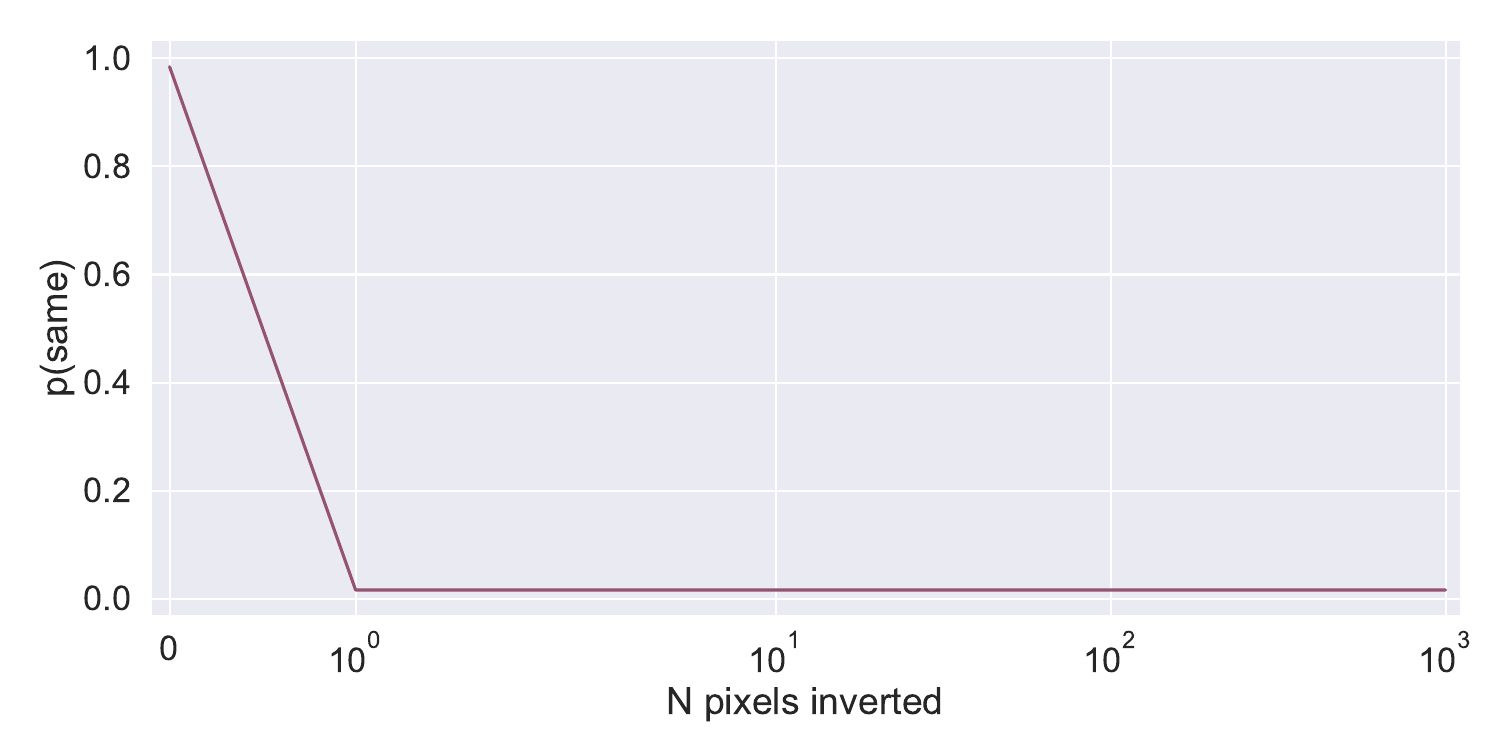}
    \caption{Random images test.} 
    \label{fig:random_imgs_test}
\end{figure}

In our second simulation we used the same 10 instances of ESBN trained on the centered SD dataset. We presented these models with pairs of random noise images, where the first image was the base and the second one was a copy of the first with a given number of pixels inverted. We generated a 1001 conditions of 100 pairs of images, from 0 to 1000 pixels inverted (see Figure \ref{fig:random_imgs}). For each of these pairs we recorded the models' predicted probability for the category \say{same}. As can be seen in Figure \ref{fig:random_imgs_test}, all ESBN instances went from a predicted probability of almost 1 to pairs of identical images to almost 0 for any pair that had one pixel inverted or more. These results that ESBN is not comparing the objects in the individual objects but the entire canvases.

\section{Simulation 5 training trajectories}
\label{section:Sim5Training}

\begin{figure*}[ht]
    \centering
    \includegraphics[width=\linewidth]{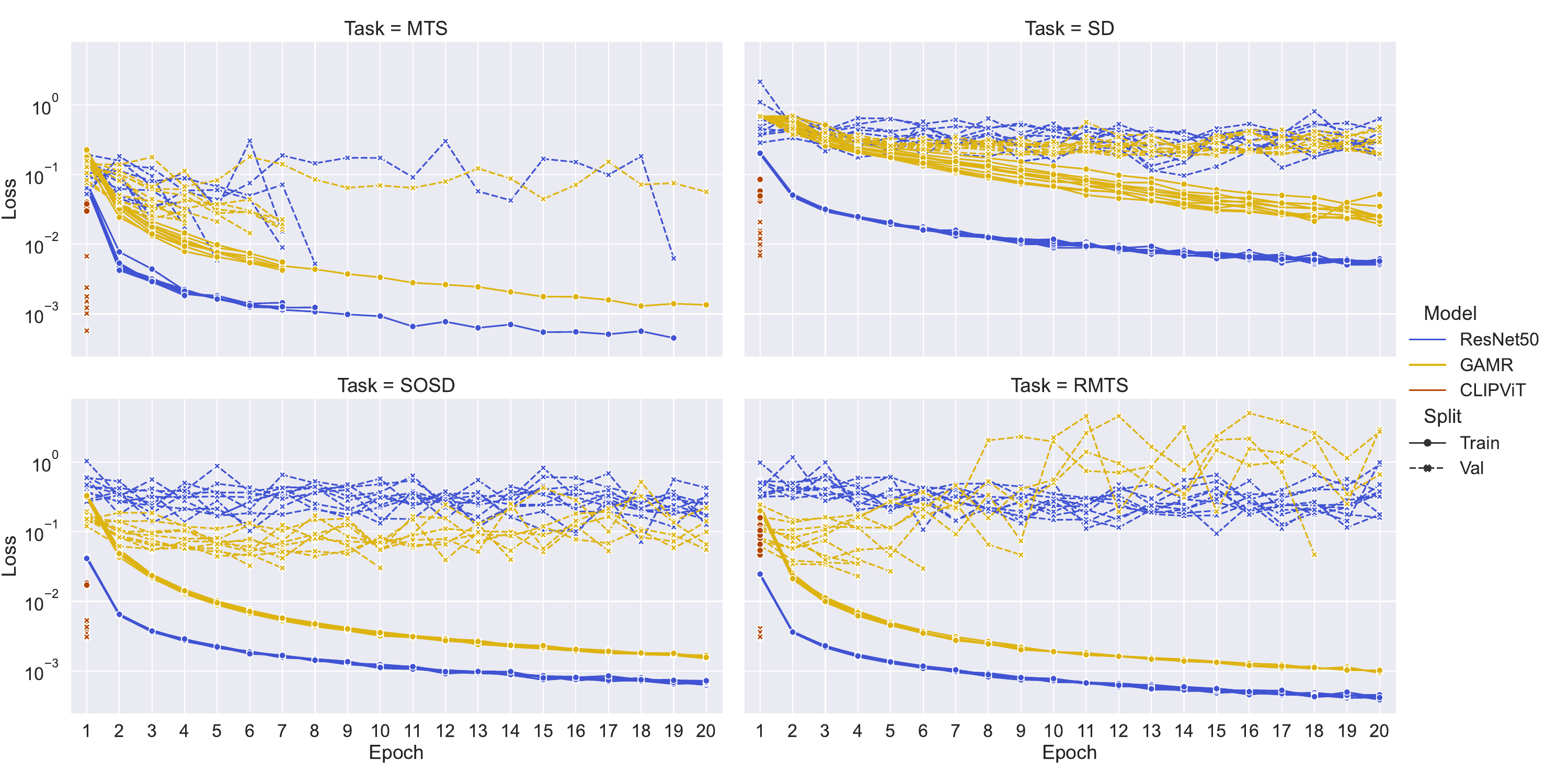}
    \caption{Training and validation loss per epoch, model and random seed in Simulation 5.}
    \label{fig:MTLtrain}
\end{figure*}

Figure \ref{fig:MTLtrain} shows the training and validations losses for all the models tested in Simulation 5. As can be seen all model instances showed a non-decreasing validation loss curve. Note that in all tasks CLIP-ViT achieve the early stop criterion of $99.0\%$ validation accuracy on the first epoch.

\printcredits

\bibliographystyle{cas-model2-names}

\bibliography{cas-refs}





\end{document}